\documentclass[10pt,twocolumn,letterpaper]{article}

\usepackage[pagenumbers]{cvpr} 
\usepackage{colortbl}  

\definecolor{superlightgray}{gray}{0.9}

\definecolor{cvprblue}{rgb}{0.21,0.49,0.74}
\usepackage[pagebackref,breaklinks,colorlinks,allcolors=cvprblue]{hyperref}

\title{MobileI2V: Fast and High-Resolution Image-to-Video on Mobile Devices}

\author{Shuai Zhang$^{*}$, Bao Tang$^{*}$, Siyuan Yu$^{*}$, Yueting Zhu, Jingfeng Yao, Ya Zou, \\ Shanglin Yuan, Li Yu, Wenyu Liu, Xinggang Wang$^{\dagger}$ \\
Huazhong University of Science and Technology\\
}

\begin{document}

\twocolumn[{%
	\renewcommand\twocolumn[1][]{#1}%
	\maketitle
	\vspace{-25pt}
	\begin{center}
		\centering
		\includegraphics[width=1 \linewidth]{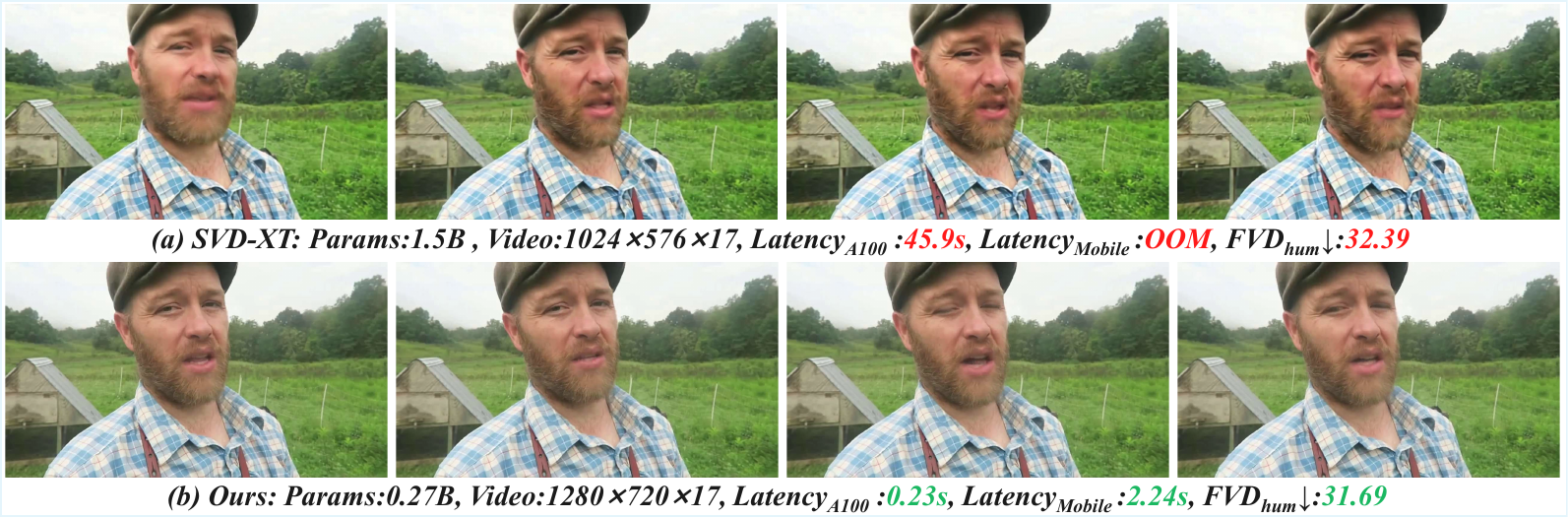}
		\label{fig:teaserfig}
		\vspace{-15pt}
		\captionof{figure}{Compared with SVD-XT ($1.5$B), our $5.55\times$ smaller MobileI2V ($0.27$B) achieves similar generation quality, using only $2.24$~s on mobile and running $199\times$ faster on an A100 GPU.
		}
	\end{center}%
	\label{teaserfig}

}]

\maketitle

\begingroup
\makeatletter
\renewcommand\@makefnmark{}
\renewcommand\thefootnote{} 
\makeatother

\footnote{
$^{*}$ Equal contribution.
}
\footnote{
$^{\dagger}$ Corresponding author (\texttt{xgwang@hust.edu.cn}).
}

\addtocounter{footnote}{-1}
\endgroup

\begin{abstract}
Recently, video generation has witnessed rapid advancements, drawing increasing attention to image-to-video (I2V) synthesis on mobile devices. However, the substantial computational complexity and slow generation speed of diffusion models pose significant challenges for real-time, high-resolution video generation on resource-constrained mobile devices. In this work, we propose MobileI2V, a 270M lightweight diffusion model for real-time image-to-video generation on mobile devices. The core lies in: (1) We analyzed the performance of linear attention modules and softmax attention modules on mobile devices, and proposed a linear hybrid architecture denoiser that balances generation efficiency and quality. 
(2) We design a time-step distillation strategy that compresses the I2V sampling steps from more than 20 to only two without significant quality loss, resulting in a 10-fold increase in generation speed. (3) We apply mobile-specific attention optimizations that yield 2× speed-up for attention operations during on-device inference. MobileI2V enables, for the first time, fast 720p image-to-video generation on mobile devices, with quality comparable to existing models. Under one-step conditions, the generation speed of each frame of 720p video is less than 100 ms. Our code is available at: https://github.com/hustvl/MobileI2V.
\vspace{-10pt}
\end{abstract}

\section{Introduction}
\label{sec:intro}
With the development of video generation models~\citep{brooks2024video,zheng2024opensora,zhou2023magicvideo,blattmann2023stable,lin2024opensoraplan}, the demand for image-to-video (I2V) generation on mobile devices is constantly increasing. I2V on mobile devices can significantly enrich system applications, such as live wallpapers. Compared to cloud-based inference, on-device inference offers distinct advantages: it eliminates dependency on network connectivity, significantly cuts data transmission and latency, and better safeguards user privacy.

\begin{figure*}[ht]
	\begin{center}
	{
		\begin{minipage}{5.65cm}
			\centering
			\includegraphics[width=1\textwidth]{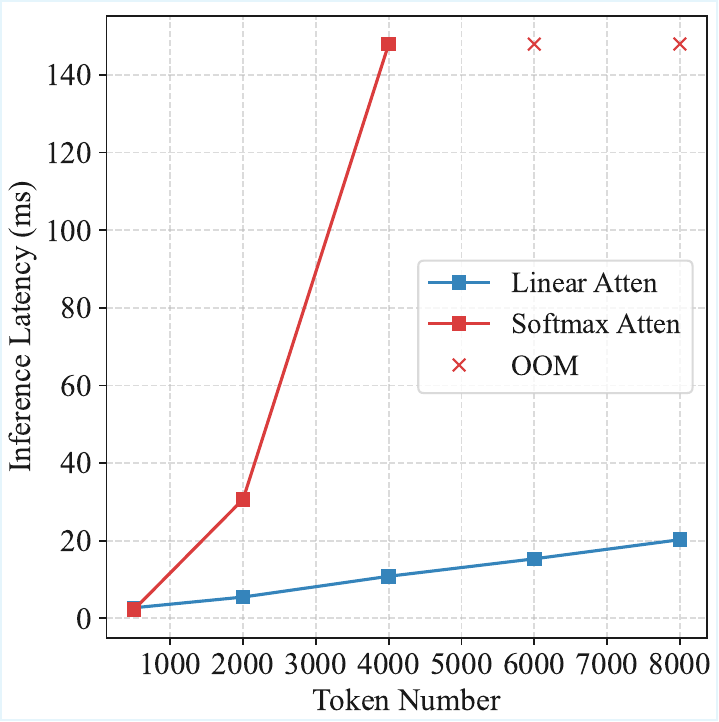}
		\end{minipage}
	}
	{
		\begin{minipage}{5.65cm}
			\centering
			\includegraphics[width=1\textwidth]{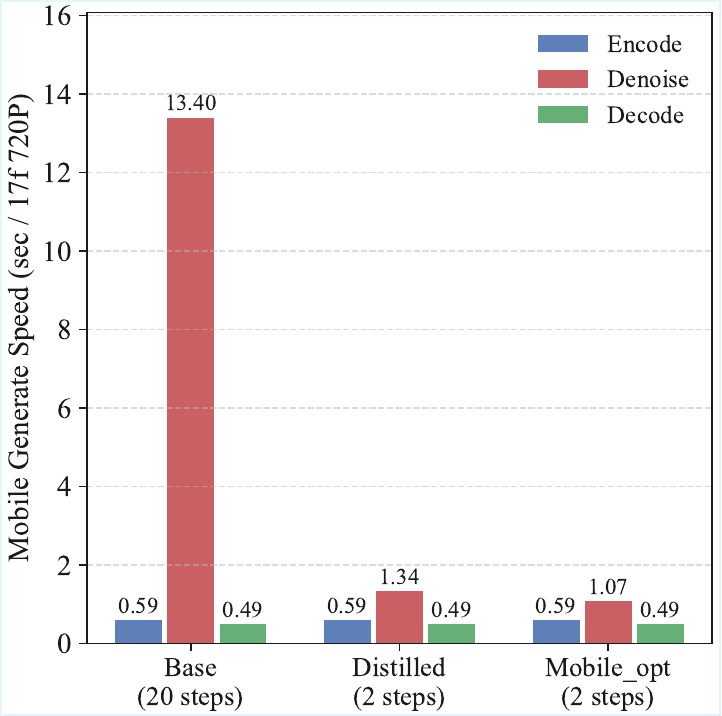}
		\end{minipage}
	}
	{
		\begin{minipage}{5.65cm}
			\centering
			\includegraphics[width=1\textwidth]{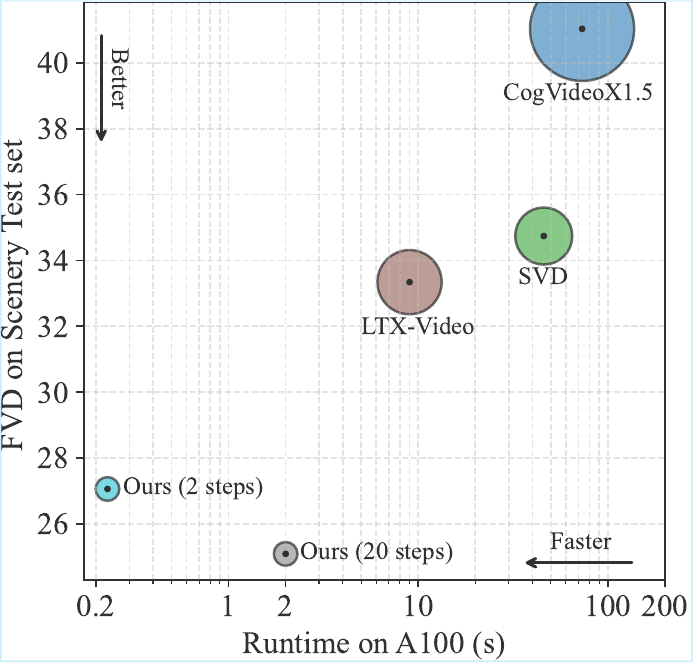}
		\end{minipage}
	}
	\end{center}
	\vspace{-18pt}
	\caption{Softmax Attention vs Linear Attention speed in mobile devices (Left). The I2V generation speed of the proposed model (Middle). Comparison results of existing I2V models (Right).} 
	\label{fig:mobilespeed}  
	\vspace{-15pt}
\end{figure*}
	
The core challenge of mobile I2V is that current diffusion models have high computational costs and slow inference speeds, which makes most models take a very long time to infer or cannot infer at all on mobile devices. The main reasons are: (1) Even after VAE compression, high-resolution videos still feed the denoiser a vast number of latent tokens. (2) The denoiser employs a DiT~\citep{PeeblesX23} built on softmax attention~\cite{VaswaniSPUJGKP17}, whose $O(n^2)$ complexity causes computation to explode as the number of tokens grows. (3) Diffusion models require multi-step reasoning when generating videos.
These reasons limit the application of video diffusion models, especially in resource-limited mobile scenarios. Therefore, this paper attempts to solve these problems and proposes a video generation model that quickly  generates high-resolution videos on mobile devices.

VAE with a high compression rate significantly reduces the number of tokens. LDM~\citep{RombachBLEO22} proposes for the first time to compress images using VAE and perform the diffusion process in the latent space. DCAE~\citep{ChenCCXYTL025} proposes a VAE with 32$\times$32 spatial down-sampling. With the high sampling rates of VAE, SANA~\citep{XieCCCT0ZLZ0025} achieves fast generation of high-resolution images. LTX-Video~\citep{hacohen2024ltx} further extended DCAE to the video generation domain, proposing a 32$\times$32$\times$8 spatial-temporal down-sampling rate VAE, which significantly improved the inference speed of T2V/I2V tasks. Therefore, we adopted this high compression ratio VAE and replaced the decoder with a lighter model~\citep{Turbo-VAED}.

Lightweight denoisers with linear complexity can result in faster inference speeds. As shown in Fig.~\ref{fig:mobilespeed}, we found that the linear attention~\citep{KatharopoulosV020} has a significant advantage over softmax attention in terms of inference speed on mobile devices. As the length of the sequence increases, the performance gap between linear attention and softmax attention becomes increasingly apparent, especially on mobile devices. Adopting a fully linear attention architecture can significantly enhance the speed of the model, but it may also reduce its accuracy. Therefore, we designed a hybrid architecture that balances the speed and accuracy of the model.

Time-step distillation significantly reduces the number of inference steps.
Currently, the inference of common video generation models requires around 20 to 50 sampling steps. After time-step distillation, the inference steps can be reduced to only 1-4 steps. Existing approaches such as LADD~\citep{LADD}, LCM~\citep{LCM}, DMD~\citep{DMD} and SF-V~\citep{SF-V} have been applied to text-to-image and text-to-video generation. We further employed time-step distillation to reduce the I2V model's inference steps from 20 to 2.

To this end, we propose MobileI2V, an I2V model that can perform fast inference on mobile devices. We use linear hybrid attention to construct a denoiser with only 270M parameters. 
As shown in Fig.~\ref{fig:mobilespeed}, after optimization, the proposed model can generate 17 frames of 720p video within 2 seconds on the iPhone 16 PM. Our contributions are as follows:

1. We propose a hybrid architecture diffusion model with only 270M parameters for 720P mobile I2V tasks. We combined the advantages of linear attention and softmax attention to balance the speed and quality of device-side generation, and for the first time explored the efficiency performance of linear attention in device-side video generation.

2. We propose a composite timestep distillation scheme for lightweight I2V models, which reduces the number of inference steps from 20 to 1–2 while maintaining comparable generative quality. This enables over 10$\times$ acceleration on mobile devices.

3. On mobile devices, we introduce three inference optimizations for linear attention: 4D channels-first layout and operator lowering, head tiling, and reduced data movement. This improves the inference speed of linear attention and softmax attention mechanisms by 2$\times$ on mobile devices.

4. To the best of our knowledge, we have achieved high-resolution (720P) I2V on a mobile device for the first time. Under one-step inference conditions, the generation time for a single frame at 720P is less than 100 ms.
\vspace{-5pt}

\section{Related Work}
\label{sec:formatting}

\paragraph{Video Diffusion Model.}
The development of video generation models is advancing rapidly~\citep{kong2025hunyuanvideo,wan2025wan,YangCogVideoX}. DiT~\citep{PeeblesX23} first introduced the Vision Transformer into diffusion models, demonstrating the scalability of Transformers for image/video generation tasks.  Latte~\citep{ma2025latte} experimented with various Transformer variants for video generation, and subsequent experiments validated the optimal Transformer variant. Opensora~\citep{zheng2024opensora} and Opensora-Plan~\citep{lin2024opensoraplan}, as outstanding open-source models within the community, have garnered widespread attention. CogVideo~\citep{hong2022cogvideo} fine-tunes pre-trained text-to-image generation models, avoiding expensive pre-training. There are currently some acceleration methods~\citep{yao2024fasterdit,yao2025reconstruction} for DiT, but generating videos on mobile devices still requires some powerful lightweight designs.
\vspace{-10pt}

\paragraph{Linear Generative Models.}
The linear attention mechanism~\citep{han2023flatten,han2024agent} has been widely attempted to be applied to generative tasks, such as image super-resolution~\citep{LinearSR}, image generation~\citep{zhu2025dig,XieCCCT0ZLZ0025}, and video generation~\citep{chen2025Sana_video,gao2024matten, huang2025linvideo}, due to its $O(n)$ computational complexity. Arflow~\citep{hui2025arflow} applies a hybrid linear model to an autoregressive flow model to improve image generation efficiency. SANA-Video~\citep{chen2025Sana_video} significantly improves video generation efficiency by utilizing Block Linear Diffusion.
In this paper, we will explore the enormous potential of linear attention mechanisms for mobile generation. 
\vspace{-10pt}

\paragraph{Mobile-Side Generative Models.}
MobileDiffusion~\citep{ZhaoMobileDiffusion} can generate high-quality 512$\times$512 images within 0.2 seconds on iPhone. MobileVD~\citep{yahia2024mobile} generating latents for a 14$\times$512$\times$256 px clip in 1.7 seconds on a Xiaomi-14 Pro. SnapGen~\citep{chen2025snapgen} demonstrates the generation of $1024^{2}$ px images on a mobile device around 1.4 seconds. SnapGen-V~\citep{wu2025snapgen} has only 0.6M parameters and can generate a 5-second video on an iPhone 16 PM within 5 seconds. These methods all employ the U-Net architecture,and the resolution of generative models on mobile devices is relatively low. In this paper, we have achieved for the first time the implementation of a 720P I2V task on a mobile device.
\vspace{-10pt}

\paragraph{Time-step Reduction.}
Time-step distillation compresses a teacher model’s multi-step trajectory into a student model that needs only 1–4 steps. Representative techniques include:
Simple direct distillation~\citep{Progressive_Distillation, CFG_Distillation} that regress the T-step chain into a one-step mapping;
Consistency-based methods (e.g., CM~\citep{Consistency_Models}, LCM~\citep{LCM}, sCM~\citep{sCM}) that enforce self-consistency along the probability-flow ODE;
Adversarial distillation (e.g., ADD~\citep{ADD}, LADD~\citep{LADD}) that uses a discriminator to align the one-step output with the teacher distribution in pixel or latent space;
VSD-based variants (e.g., DMD~\citep{DMD}, DMD2~\citep{DMD2}, SiD~\citep{SID}, SIM~\citep{SIM}) that match distributions via variational score distillation while suppressing mode collapse.
Most of these methods are used for text-to-image tasks, and recent work such as SF-V~\citep{SF-V} is transferring them to video generation by adding spatio-temporal discriminators to maintain frame-wise consistency.
\vspace{-5pt}

\section{Preliminaries}

\paragraph{Flow-based Diffusion Models.}
The input video data is $V\in \mathbb{R}^{H\times W\times 3\times T}$, where H, W, and T represent the length, width, and number of frames of the video, respectively. Compress to latent $L\in \mathbb{R}^{\hat{H}\times \hat{W}\times 3\times \hat{T}}$ via VAE, where $\hat{H}=H/32$,$\hat{W}=W/32$ and $\hat{T}=T/8$. The noise $N\in \mathbb{R}^{\hat{H}\times \hat{W}\times 3\times \hat{T}}$ is denoised by a denoiser and then decoded into video by a decoder. We use the SD3~\citep{EsserKBEMSLLSBP24} optimizer to train and infer our model. We define a forward process, corresponding to a probability path between $p_0$ and $p_1$ = $\mathcal N (0, 1)$, as
\begin{equation}
	z_t = a_{t}x_0 + b_{t}\epsilon, \ \ where\  \epsilon\in \mathcal N(0,I).
\end{equation}

The objective function of the model is:
\begin{equation}
    L(x_0)=\mathbb{E}_{t,\epsilon}[w_t\lambda_{t}^{\prime}||\epsilon(z_t,t)-\epsilon||^2,
\end{equation}
where $w_t=-\frac{1}{2}\lambda_t^{\prime}b_{t}^{2}$,  $\lambda_{t}=log\frac{a_t^2}{b_t^2}$ and $\lambda_t^{\prime}=2(\frac{a_t^{\prime}}{a_t}-\frac{b_t^{\prime}}{b_t})$.

Rectified Flows~\citep{LiuG023} defines the forward process as a straight line path between the data distribution and the standard normal distribution. So for it,  $a_{t}=1-t$, $b_{t}=t$.

\vspace{-10pt}
\paragraph{Softmax Attention and Linear Attention.}
Given tokens of length $N$, the self attention calculation method is as follows~\cite{HanPHSH23}:

\begin{equation}
    O_{i}=\sum_{j=1}^{N}\frac{\mathrm{Sim}(Q_{i},K_{j})}{\sum_{j=1}^{N}\mathrm{Sim}(Q_{i},K_{j})}V_{j},
\end{equation}
where $Q=xW_{Q}$, $K=xW_{K}$, $V=xW_{V}$. $W_{Q}$, $W_{K}$, $W_{V}\in\mathbb{R}^{C \times C}$ are projection matrices and $Sim(,)$ denotes the similarity function. For Softmax attention, $Sim(Q,K)=exp(QK^T/\sqrt{d})$.
For Linear Attention, we refer to the approach used in SANA~\citep{XieCCCT0ZLZ0025}, $\mathrm{Sim}(Q,K)=\mathrm{ReLU}(Q_{i})\mathrm{ReLU}(K_{j})^T$.

\section{Method}

\begin{figure*}[t]
	\centering
	\includegraphics[width=1\textwidth]{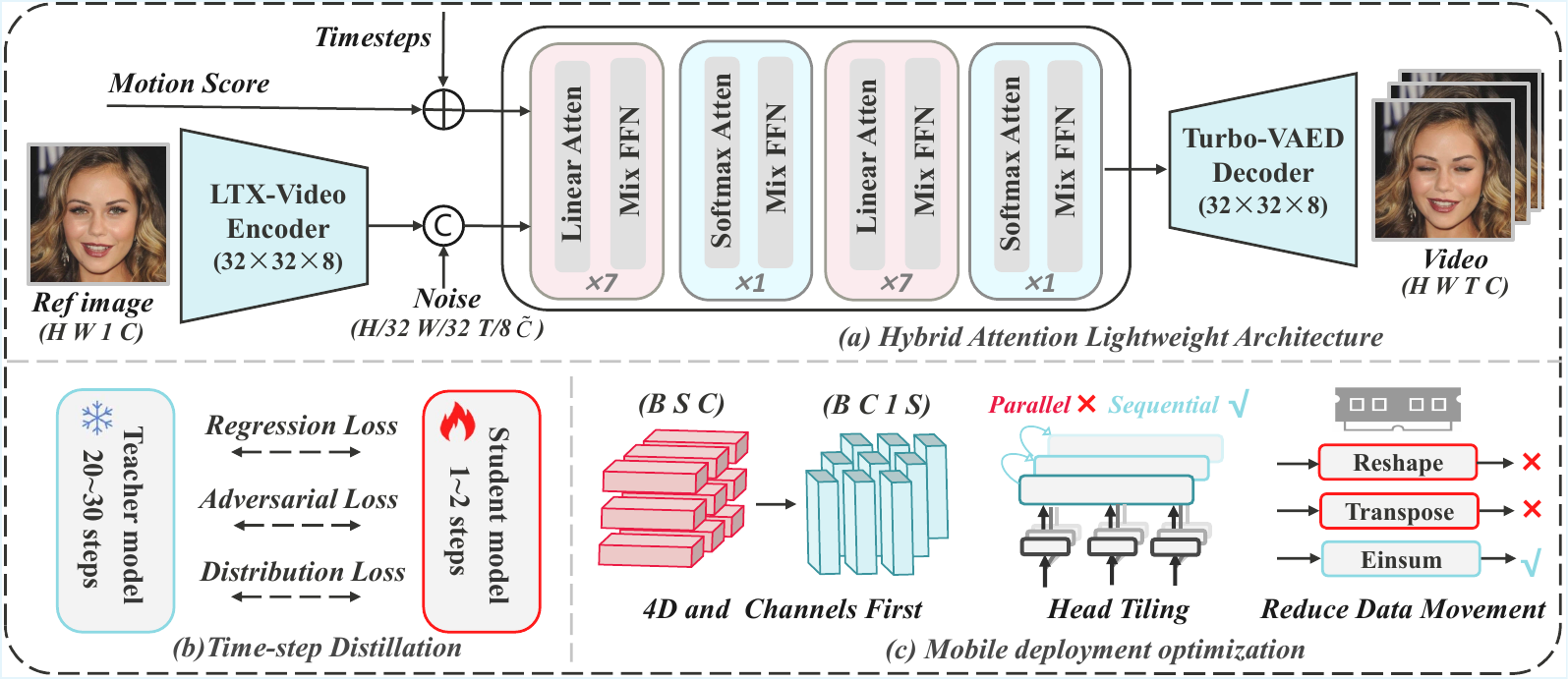}
	\vspace{-15pt}
	\caption{Our proposed MobileI2V framework employs a hybrid attention framework to support fast inference, utilizes time-step distillation to reduce the number of inference steps, and finally optimizes the model on mobile devices to further accelerate inference.}
	\label{fig:framework}
	\vspace{-15pt}
\end{figure*}

\subsection{Model Architecture}

Our framework diagram is shown in Fig.~\ref{fig:framework}. We design a hybrid linear DiT as a noise reducer and use a 32$\times$32$\times$8 high compression ratio VAE to reduce the number of tokens.

\vspace{-10pt}
\paragraph{Hybrid Linaer DiT.}
We follow SANA~\citep{XieCCCT0ZLZ0025}'s architecture. We used 16 layers of Dit, with only 0.27M parameters. Due to the $O(n^2)$ complexity of the ordinary Softmax Attention, the complexity increases quadratically when processing high resolution. As shown in Fig.~\ref{fig:mobilespeed}, the performance of linear attention is significantly faster than the traditional vanilla attention on mobile devices. Therefore, in this paper, we replaced part of the softmax attention with linear attention. We drew inspiration from the hybrid architecture in MiniMax-01~\citep{minimax2025}, where a softmax attention follows every seven linear attentions. The experiment found that introducing two layers of softmax attention can significantly compensate for the accuracy loss caused by linear attention. Therefore, we adopted a hybrid architecture to balance the model's performance and accuracy. We added positional encoding to the model based on Allegro~\citep{zhou2024allegroopenblackbox}.

\vspace{-10pt}
\paragraph{High Compression Ratio VAE.}
We use the LTX-Video~\citep{hacohen2024ltx} VAE encoder and the Turbo-VAED~\citep{Turbo-VAED} decoder, with a downsampling factor of 32$\times$32$\times$8. The number of channels is 128. For a 720p video, the latent size after VAE is 40$\times$23$\times$3. Turbo-VAED distills the VAE of LTX-Video, reducing the size of the model decoder and significantly improving decoding speed. The decoding speed of the distilled VAE is 3$\times$ that of the LTX-Video VAE.

\subsection{Design of Image to Video}
The first frame of the video is used as the reference image in the I2V task. As shown in Fig.~\ref{fig:framework}, the reference image, after VAE encoding, is used to replace the first frame in the noisy latent. We refer to the design of LTX-Video~\citep{hacohen2024ltx}, setting time $t$ as independent for each token and setting the reference frame's $t$ to 0. We extracted the optical flow scores from the training data and used them as conditions to be fed into denoiser, thereby controlling the motion magnitude of the objects. Conditioned on a motion score, the I2V model converges faster and allows users to intuitively control the generated video’s motion intensity.


\subsection{Time-step Distillation}

\begin{figure*}[th]
	\centering
	\includegraphics[width=1\textwidth]{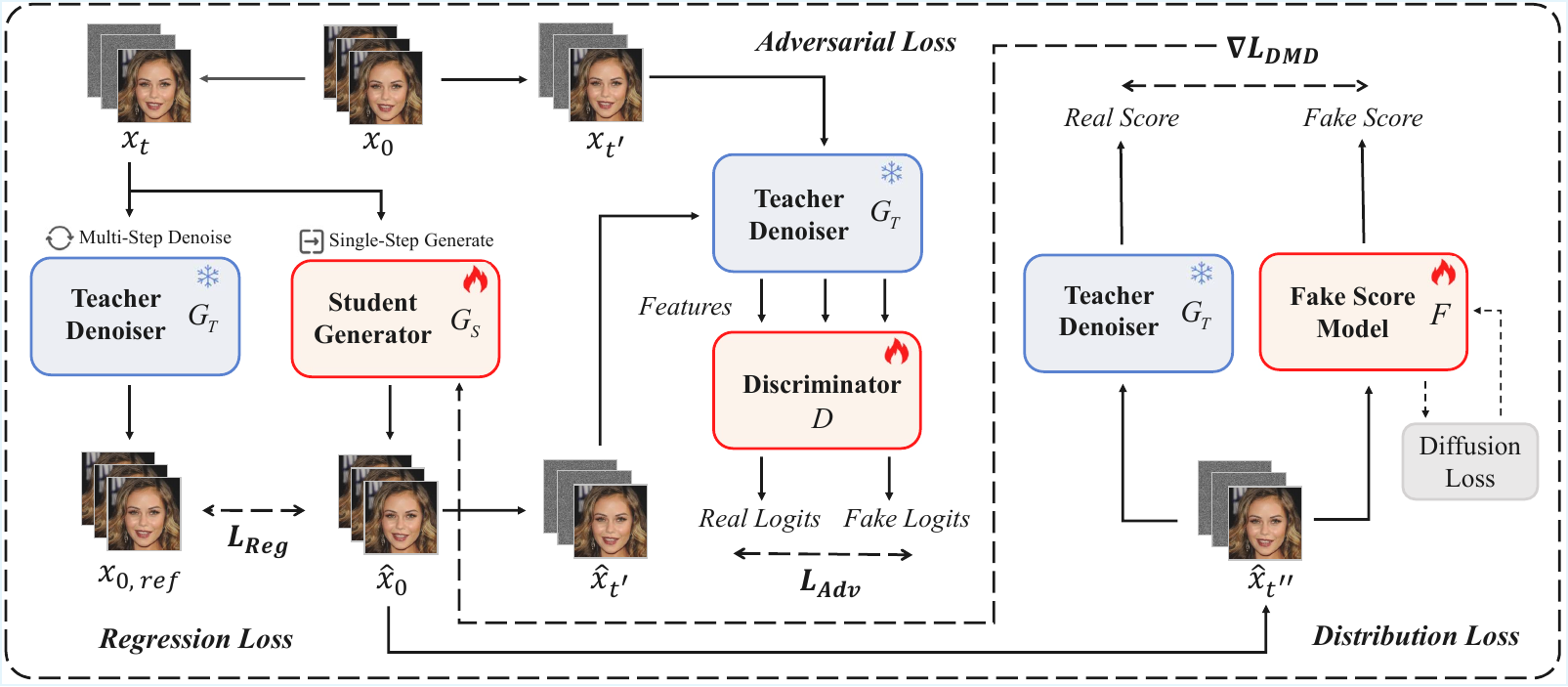}
	\vspace{-15pt}
	\caption{The proposed i2v time-step distillation framework. It includes three parts: regression loss, adversarial loss, and distribution loss.}
	\label{fig:ad}
	\vspace{-15pt}
\end{figure*}

As shown in Fig.~\ref{fig:ad}, we explore a timestep distillation strategy that combines multiple techniques, enabling video generation within only one or two inference steps. Specifically, inspired by successful practices in text-to-image (T2I) distillation~\citep{DMD,LADD,Flash_Diffusion}, we employ three complementary losses: regression loss, adversarial loss, and distribution matching loss. To this end, the distillation process jointly maintains a Student Model \(G_S\), a Teacher Model \(G_T\), a Fake Score Model \(F\), and a Discriminator \(D\). Among them, \(G_S\) , \(G_T\), and \(F\) are initialized from pretrained weights, while \(D\) is randomly initialized. During training, the parameters of \(G_S\), \(F\), and \(D\) (denoted as \(\theta_S\), \(\theta_F\), and \(\psi\), respectively) are optimized, whereas the teacher \(G_T\) remains frozen. The three losses are designed as follows:

\vspace{-15pt}
\paragraph{Regression Loss.}
A clean video \(x_0\) is perturbed to a noisy state \(x_t\) at a given noise level. Both the student model \(G_S\) (single-step generation) and the teacher model \(G_T\) (multi-step generation) are applied to \(x_t\). The mean squared error (MSE) between their predictions is used as the regression loss:

\vspace{-15pt}
\begin{equation}
    \mathcal{L}_{\mathrm{reg}}(\theta_S) 
    = \mathbb{E}_{x_0,\, t,\, \epsilon} 
    \Big[
    \big\| G_S(x_t, t) - \hat{x}_0^{\mathrm{Teacher}} \big\|_2^2
    \Big].
\end{equation}

Here, $\hat{x}_0^{\mathrm{Teacher}}$ denotes $G_T$'s multi-step generated video starting from $x_t$.

\vspace{-15pt}
\paragraph{Adversarial Loss.}
The predicted clean video \(\hat{x}_0\) from \(G_S\) and the ground-truth clean video \(x_0\) are treated as adversarial pairs. After adding noise of a certain magnitude, they are passed into the frozen teacher \(G_T\) to extract features, which are subsequently fed into the discriminator \(D\). The adversarial loss is then computed as:

\vspace{-15pt}
\begin{equation}
    \begin{aligned}
        \mathcal{L}_{\mathrm{adv}}^{G}(\theta_S) 
        &= -\, \mathbb{E}_{x_{0},\, t,\, \epsilon} 
        \Bigg[
        \sum_{k} D_{\psi,k} \Big( 
        G_T^{(k)}(\hat{x}_t, t) 
        \Big)
        \Bigg], \\[2mm]
        \mathcal{L}_{\mathrm{adv}}^{D}(\psi) 
        &= \mathbb{E}_{x_{0},\, t,\, \epsilon} 
        \Bigg[
        \sum_{k} \mathrm{ReLU}\!\Big(
        1 - D_{\psi,k}\big( 
        G_T^{(k)}(x_t, t) 
        \big)
        \Big) \\[1mm]
        &\quad +\,
        \sum_{k} \mathrm{ReLU}\!\Big(
        1 + D_{\psi,k}\big( 
        G_T^{(k)}(\hat{x}_t, t) 
        \big)
        \Big)
        \Bigg].
    \end{aligned}
\end{equation}

Here, $x_t$ and $\hat{x}_t$ are obtained by adding noise to $x_0$ and $\hat{x}_0$, respectively. $G_T^{(k)}(\cdot)$ denotes $G_T$'s output at the $k$-th feature branch.

\vspace{-15pt}
\paragraph{Distribution Matching Loss.}
The predicted clean video \(\hat{x}_0\) from \(G_S\) is first perturbed by noise. The resulting noisy sample is evaluated by both \(G_T\) and the fake score model \(F\), producing a real score and a fake score, respectively. Following the formulation in DMD~\citep{DMD}, we compute the distribution matching loss as:
\vspace{-5pt}
\begin{equation}
    \begin{aligned}
        & \nabla \mathcal{L}_{\mathrm{DM}}(\theta_S) 
        = \mathbb{E}_{t} \Big[ \nabla_{\theta_S} \mathrm{KL}\big( p_{\mathrm{fake}, t} \,\|\, p_{\mathrm{real}, t} \big) \Big] \\
        &= - \, \mathbb{E}_{t} \Bigg[
        \int 
        \Big( s_{\mathrm{r}}(\hat{x}_t, t) 
        - s_{\mathrm{f}}(\hat{x}_t, t) \Big)
        \frac{dG_S(x_t, t)}{d\theta_S} \, dx_t
        \Bigg],
    \end{aligned}
\end{equation}

\begin{equation}
    \mathcal{L}_{\mathrm{DM}}(\theta_F) 
    = \mathbb{E}_{\hat{x}_0,\, t,\, \epsilon} 
    \Big[
    \big\| F(\hat{x}_t, t) - \hat{x}_0 \big\|_2^2
    \Big].
\end{equation}

Here, $s_{\mathrm{r}}$ and $s_{\mathrm{f}}$ denote the scores computed by $G_T$ and $F$, respectively. The noisy sample $\hat{x}_t$ is obtained by adding noise to the  $\hat{x}_0$.


\subsection{iPhone Mobile Model Deployment}

Following Apple’s guidelines in \emph{Deploying Transformers on the Apple Neural Engine (ANE)}, we optimize all attention computations (softmax attention and linear attention) by aligning data layout and operator mappings with the ANE compiler’s preferences. 
We apply three families of optimizations:

\vspace{-15pt}
\paragraph{4D and Channels-First.}
We represent intermediate tensors as (B, C, 1, S) and keep standard Linear projections. After projection we reshape with lightweight transpose so Q/K/V live in a 4D channels-first layout during attention. The ANE keeps the last axis unpacked: when sequence length occupies that axis, accesses along it are contiguous, improving prefetch efficiency and minimizing the impact of padding. In contrast, making a channel dimension the last axis inflates buffers and hurts L2 residency.

\vspace{-15pt}
\paragraph{Head Tiling.}
When computing on a GPU, we compute multi-head attention in parallel. However, when computing on a mobile device, we compute each head individually and serially. Multi-head attention is explicitly split into per-head $Q$, $K$, and $V$, computed on smaller tiles using a per-head function list, improving L2 locality and post-compilation multicore utilization.

\vspace{-15pt}
\paragraph{Reduced Data Movement.}
Along the attention path, redundant \texttt{reshape}/\texttt{transpose} operations are eliminated, retaining only a single necessary transpose on $K$. The core products are implemented using an \texttt{einsum} form that maps directly to batched matrix multiplies on hardware, substantially lowering bandwidth pressure. 

\begin{figure*}[h!]
    \centering
    \includegraphics[width=1\linewidth]{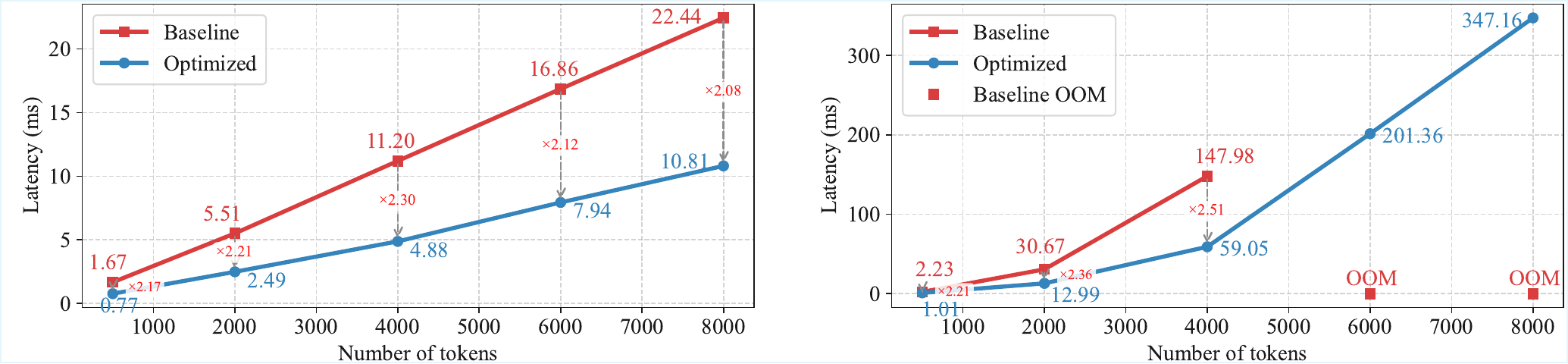}
    \vspace{-15pt}
    \caption{On-device latency on iPhone 16 Pro versus sequence length for linear attention (left) and softmax attention (right).} 
    \vspace{-10pt}
    \label{fig:deployment1}
\end{figure*}

As show in Fig~\ref{fig:deployment1}, these optimizations speed up attention computation by more than 2$\times$ for both linear and softmax attention modules. This further improves the inference speed of the model.

\vspace{-5pt}

\begin{figure*}[h!]
	\centering
	\includegraphics[width=1\linewidth]{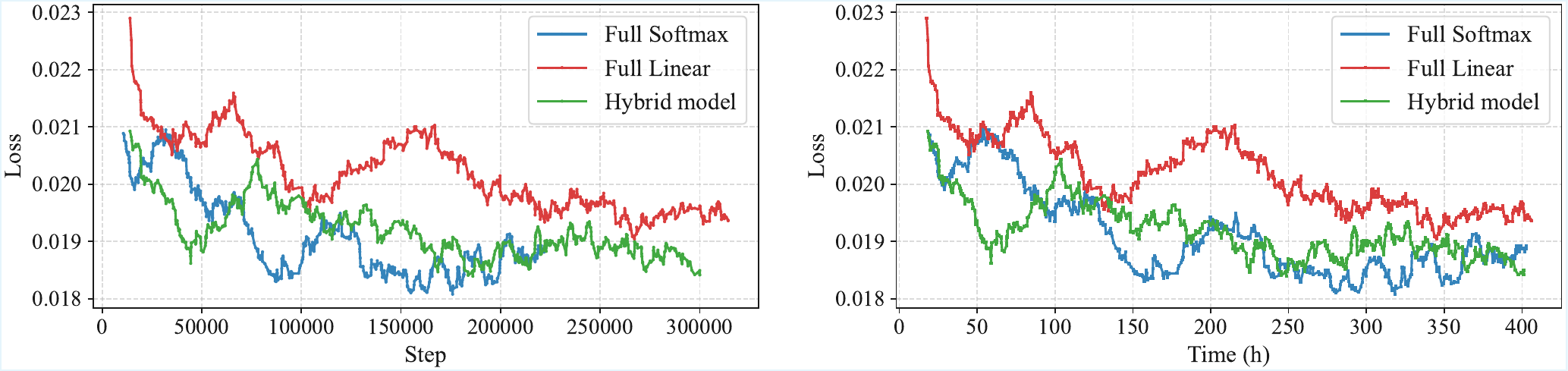}
	\vspace{-20pt}
	\caption{Loss curves for training steps (left) and training time (right) for different attention models.} 
\vspace{-5pt}
\label{fig:4_2}
\end{figure*}

\begin{table*}[h!]
	\centering
	\begin{tabular}{ccccccccc}
		\toprule
		Model  & {Type} & {Steps}  &{Params}& {Resolution} & {Latency$_{A100}$} & {Latency$_{Mobile}$}& {FVD$_{hum}$$\downarrow$} & {FVD$_{scen}$$\downarrow$}\\
		\midrule
		DynamiCrafter~\cite{xing2024dynamicrafter} &U-Net  & 30 &1.1B&  1024$\times$576 & 57.1 s& OOM&67.98&39.19 \\
		CogVideoX1.5~\cite{yang2024cogvideox} & DiT & 30 &5.0B&  720$\times$480 &73.1 s & OOM&60.83 &41.03 \\
		SVD-XT~\cite{blattmann2023stable} & U-Net & 30 &1.5B & 1024$\times$576 &45.9 s & OOM&  32.39&34.74 \\ 
		LTX-Video~\cite{hacohen2024ltx} & DiT & 30 &1.9B& 1280$\times$720&9.00 s & OOM& 36.04&33.34 \\
		Ours & DiT & 30 &0.27B& 1280$\times$720&\underline{2.00 s}  &\underline{20.13 s}  &  \textbf{26.99}   &  \textbf{25.09} \\
		Ours (Distilled) & DiT & 2 &0.27B &1280$\times$720&\textbf{0.23 s} &\textbf{2.24 s} &  \underline{31.69}&\underline{27.06}   \\
		\bottomrule
	\end{tabular}
\vspace{-5pt}
\caption{Speed and quality comparison of different models performing 17-frame I2V tasks.} 
\vspace{-10pt}
\label{table_model_comparasion}
\end{table*}

\section{Experiment}
\label{others}

\subsection{Data Process and Training}

\paragraph{Data Process.}
There are many open source video datasets, such as Openvid~\citep{NanXZFYCL0T25}, VFHQ~\citep{XieWZDS22} and Celebv-text~\citep{YuZJL0W23}. In terms of data, we collected face videos and landscape videos with resolutions of 512 and 720P. The 512 resolution data is approximately 160K (landscape 115K + face 45K), and the 720P data is approximately 170K (landscape 93K + face 88K). We filtered the data by aesthetic score, optical flow score, and reconstruction score to select a batch of high-quality data. We used DOVER~\citep{Wu0LCHWSYL23} to evaluate the aesthetic and clarity scores of the videos and filtered out data below 70. We refer to the optical flow evaluation in opensora~\citep{zheng2024opensora} and obtain videos with scores greater than 0.1 and less than 6.5. The high compression rate VAE performs poorly in reconstructing some data, so we selected data with a reconstruction PSNR greater than 32 by calculating the PSNR before and after compression.

\begin{figure*}[th]
	\centering
	\includegraphics[width=1\textwidth]{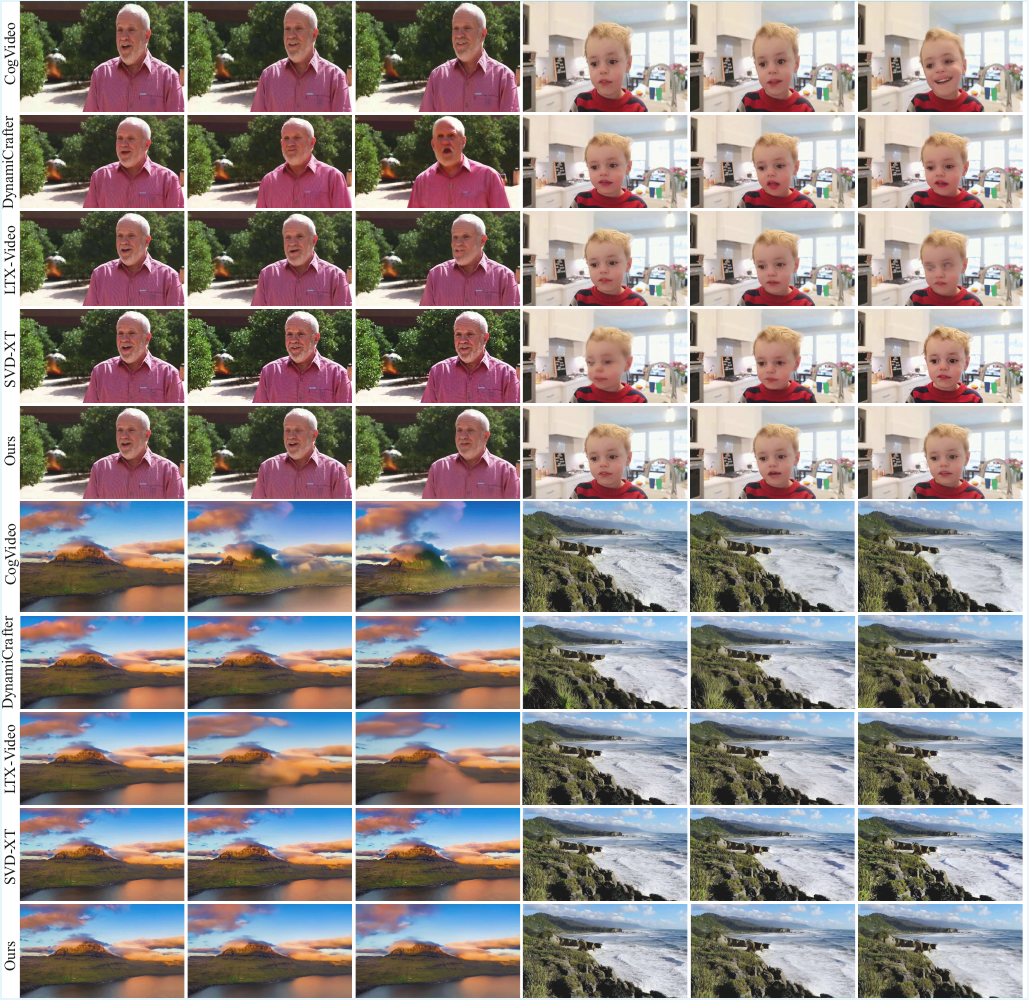}
	\vspace{-15pt}
	\caption{Visualization of 30-step Image-to-Video inference results across different models.}
	\label{fig:motivation1}
	\vspace{-15pt}
\end{figure*}

\vspace{-10pt}
\paragraph{Training.}
Training the model on 24 Nvidia V100 32G GPUs would take approximately one week. We use the CAME~\citep{LuoRZJ0023} optimizer with a learning rate of 1$e$ - 4. We first train the model using low-resolution videos, and then we train it using high-resolution videos. This helps the model converge faster.

\vspace{-5pt}

\subsection{Evaluation}
We adopt FVD~\citep{UnterthinerSKMM19} as the metric for video generation quality evaluation. We collected 1,348 facial data samples and 1,739 aerial landscape data samples to serve as the test set.

We compared the generation speeds of different models on the A100 and iPhone, as well as the generation results on two datasets, as shown in Tab.~\ref{table_model_comparasion}. The visualization results are shown in Fig.~\ref{fig:motivation1}. Existing I2V models, such as CogVideoX1.5, DynamiCrafter, SVD, and LTX-video, cannot run on mobile devices due to their huge number of parameters. 
It can be observed that our method is significantly faster than existing I2V methods, is deployable on mobile devices, and achieves comparable performance to current I2V methods in both facial and landscape scenarios. After distillation, the generation speed of the two-step model is significantly improved, and there is no obvious decrease in the indicators.
\section{Ablation Experiment}

\vspace{-5pt}
\paragraph{Hybrid Architecture.}
In order to verify the effectiveness of the proposed hybrid architecture model, we compared the hybrid architecture model with the full linear attention and full softmax attention models. As can be seen from the Fig.~\ref{fig:4_2} and Tab.~\ref{Hybrid_Architecture}, the loss reduction curve of the hybrid model with only two attention layers is close to that of the full softmax model. The full softmax model demonstrates strong representational capabilities, as evidenced by its best performance in the FVD$_{scen}$ metric. However, its speed is much slower than the full linear model and the hybrid model. The hybrid model is slightly slower than the full linear model in terms of speed, but outperforms the full linear model in terms of metrics.
\vspace{-10pt}

\begin{table}[h]
\centering
\resizebox{\linewidth}{!}{
\begin{tabular}{cccccc}
\toprule
Model   & {FVD$_{hum}$$\downarrow$} & {FVD$_{scen}$$\downarrow$} & {Speed$_{v100}$$\downarrow$} &{Speed$_{mobile}$$\downarrow$}\\
\midrule
Linear    &   27.47  & 25.75 & 3.72s  &13.74s    \\
Softmax     &   28.89  & 23.66 & 7.88s  &22.97s   \\
Hybrid     &   26.99   &  25.09& 4.64s &14.85s \\
\bottomrule
\end{tabular}}
\vspace{-5pt}
\caption{ Ablation study on the effect of hybrid architecture. }
\vspace{-10pt}
\label{Hybrid_Architecture}
\end{table}

\begin{figure*}[h]
\centering
\includegraphics[width=1\textwidth]{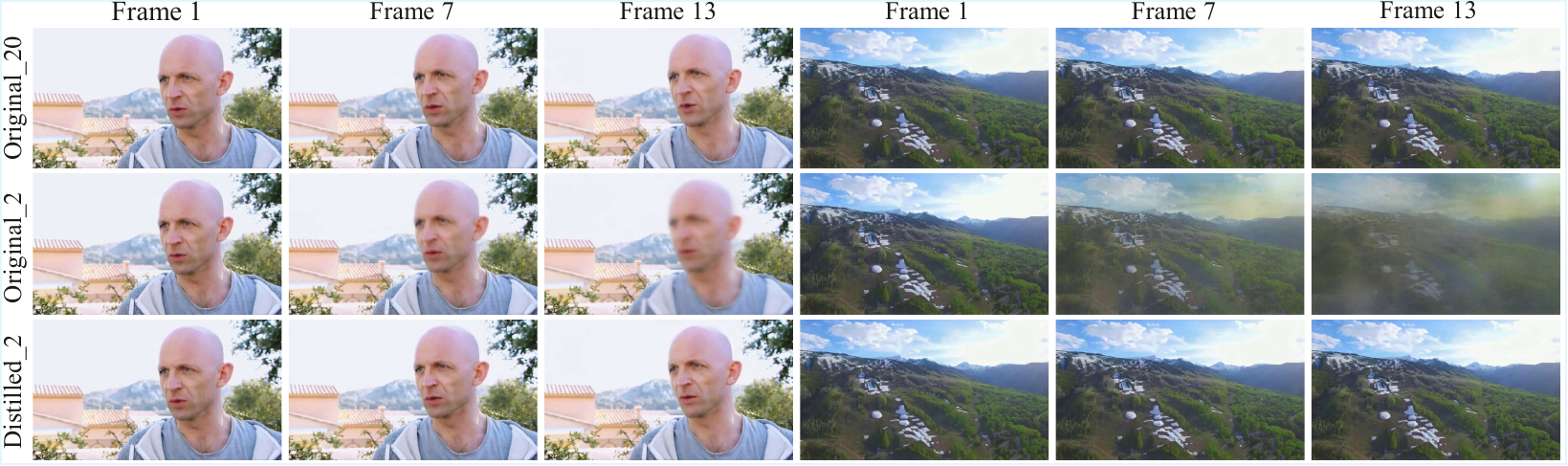}
\vspace{-15pt}
\caption{Visualization results of time-step distillation ablation experiment. Original$\_$20: 20-step results of the undistilled model. Distilled$\_$2: 2-step results of the model after distillation. The first frame is the reference image input to the model.}
\vspace{-10pt}
\label{fig:distill_ablation}

\end{figure*}

\vspace{-10pt}
\paragraph{Time-step Distillation.}
To validate the effectiveness of timestep distillation, we compare two-step inference outputs of the model before and after distillation, as illustrated in Fig.~\ref{fig:distill_ablation}. 
In videos generated with only 2 inference steps, the non-distilled model exhibits noticeable blurring in later frames, while the distilled model effectively eliminates this blurring and produces outputs that closely approximate those obtained with 20 steps. As shown in Tab.~\ref{distillation}, the metrics for the model’s one-step generation improve markedly after distillation.

\noindent\textbf{Distillation Loss Function.} To evaluate the contributions of each loss in the time-step distillation, we conducted ablation studies, with the results summarized in Tab.~\ref{distillation}. The key findings are as follows:
(1) Enabling QK normalization in all attention layers and applying timestep normalization significantly improves the visual quality of single-step generation.
(2) Adding the adversarial loss on top of the regression loss further enhances the metrics, but the performance still falls short of the full scheme. We attribute this to the limited representation capacity of the lightweight I2V backbone, which restricts the discriminator from providing sufficiently strong supervision.
(3) Incorporating only the distribution matching loss, in contrast, degrades performance. Visualization shows that although this loss increases motion magnitude, it introduces noticeable blurriness, thereby impairing overall perceptual quality.
(4) Jointly training with all three losses achieves the best results, demonstrating complementary benefits. The single-step generation output exhibits substantially improved sharpness and temporal consistency compared to any individual or pairwise combination of losses.

\vspace{-5pt}
\begin{table}[h!]
	\centering
    \resizebox{\linewidth}{!}{
	\begin{tabular}{cccccccc}
		\toprule
		Model  &  {NFE}&{Norm} & {$L_{reg}$} & {$L_{adv}$} & {$L_{dm}$} & {FVD$_{h}$$\downarrow$} & {FVD$_{s}$$\downarrow$}\\
		\midrule
		Teacher&1& - &  -  &   -    &   -  &144.85& 110.13 \\
		Teacher&1& \checkmark &  -  &   -    &   -  &122.05& 104.95 \\
		Distilled&1&   \checkmark &   \checkmark   &   -    &   -  &53.08&40.20  \\
		Distilled&1&   \checkmark &  \checkmark    &  \checkmark   &   -   &51.93& 42.32\\
		Distilled&1&   \checkmark &  \checkmark    &  -   & \checkmark &    115.45 & 56.07 \\
		Distilled&1&   \checkmark &  \checkmark    &  \checkmark   & \checkmark & 43.62   &32.68 \\
		Distilled&2&   \checkmark &  \checkmark    &  \checkmark   & \checkmark & 31.69    &27.06 \\
		\bottomrule
	\end{tabular}}
\vspace{-5pt}
\caption{ Ablation study on time-step distillation loss function. }
\vspace{-10pt}
\label{distillation}
\end{table}

\begin{figure}
	\centering
	\includegraphics[width=1\linewidth]{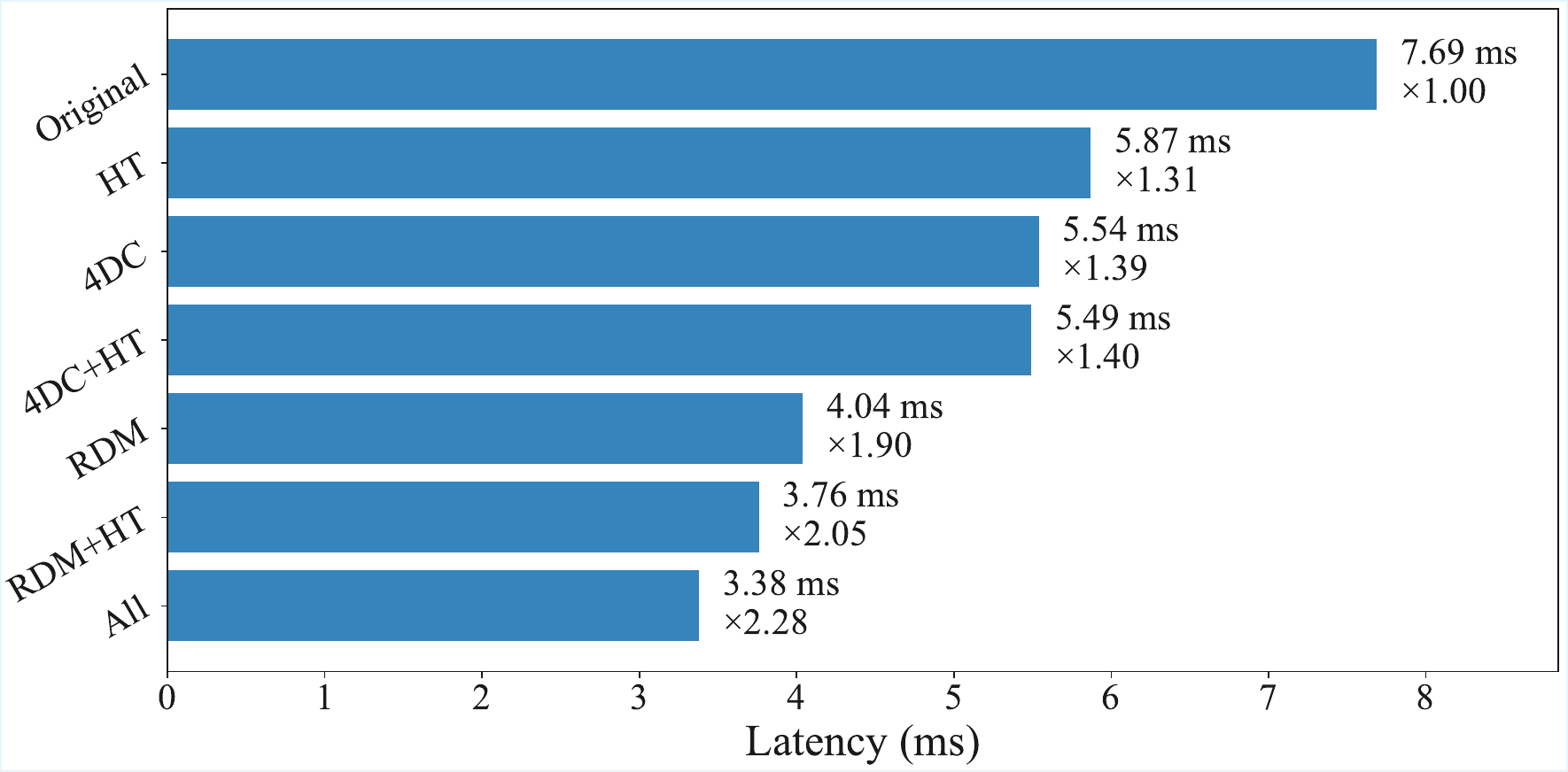}
        \vspace{-20pt}
	\caption{ Ablation study of linear attention optimizations on Core ML (iPhone 16 Pro).} 
	\label{fig:deployment2}
	\vspace{-15pt}
\end{figure}

\vspace{-10pt}
\paragraph{Mobile Optimization.}
We ablate three implementation optimizations—4DC (4D channels-first), RDM (reduced data movement), and HT (head tiling)—on Core ML (iPhone 16 Pro). The ablation study uses linear attention and processes 2,760 tokens. The results are in Fig.~\ref{fig:deployment2}. Applying all three optimizations together reduces the latency from 7.69 ms to 3.38 ms, yielding a 2.27× speed-up. Among the three optimizations, RDM yields the greatest improvement: latency drops to 4.04 ms, a 1.9× speed-up. Applying the 4DC and HT optimizations on top of RDM shaves off another 0.66 ms. The experiment demonstrates the effectiveness of all three optimizations and shows that they can be stacked for maximum gain.

\vspace{-5pt}
\section{Limitations and Future Work}
Although the proposed model demonstrates a clear speed advantage over existing ones, it still has several limitations. First, due to limited training resources and data availability, the proposed I2V model is not universal. Subsequent larger-scale data training is needed to make it more general. Furthermore, due to the high compression rate of VAE, some areas of faces and landscapes may be blurry. Increasing the motion amplitude of the video exacerbates this problem.
To address these issues, it is crucial to research higher-quality high-compression VAEs. Furthermore, further research on stronger linear attention models~\citep{Gated_Linear_Attention,gu2024mamba} and their downstream applications will be a promising research direction.

\vspace{-5pt}
\section{Conclusion}

In this paper, we propose a lightweight diffusion model for fast I2V on mobile devices. We have discovered that a hybrid architecture combining linear and softmax layers can effectively balance generation speed and quality, proving highly efficient on mobile platforms. By employing high compression rates, we significantly reduce the length of tokens and utilize time-step distillation to decrease the number of sampling steps to 1-2, thereby substantially enhancing video generation speed. In addition, we also proposed three optimization methods for the linear attention on the mobile side. Under the condition of one-step inference, the fastest mobile video generation speed can reach 96ms per frame. MobileI2V enables fast, high-resolution image-to-video generation on mobile devices, poised to power a broad range of future on-device applications.

{
    \small
    \bibliographystyle{ieeenat_fullname}
    \bibliography{main}
}

\clearpage
\setcounter{page}{1}
\maketitlesupplementary

\label{sec:rationale}
In this appendix, we provide more experimental results in Section~\ref{s9} and mobile test results in Section~\ref{s10}. Section~\ref{s11} provides the limitations of the paper and future work. 

\vspace{-5pt}
\section{More Experimental Results}
\label{s9}
\subsection{VbenchI2V Results}

\begin{figure*}[h]
	\centering
	\includegraphics[width=1\linewidth]{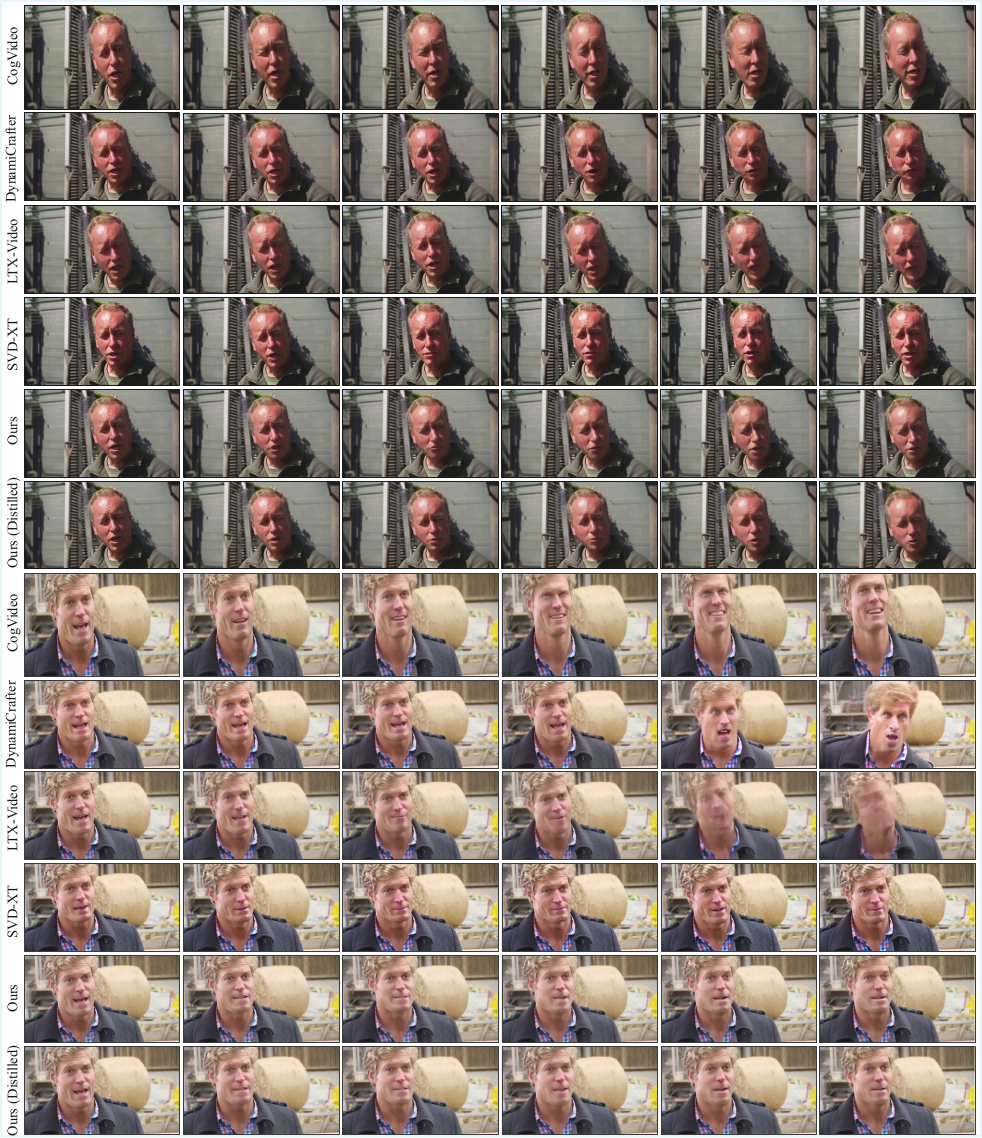}
	\caption{More visual comparison results.} 
	\label{fig:more_vis1}
	\vspace{-10pt}
\end{figure*}

\begin{figure*}[h]
	\centering
	\includegraphics[width=1\linewidth]{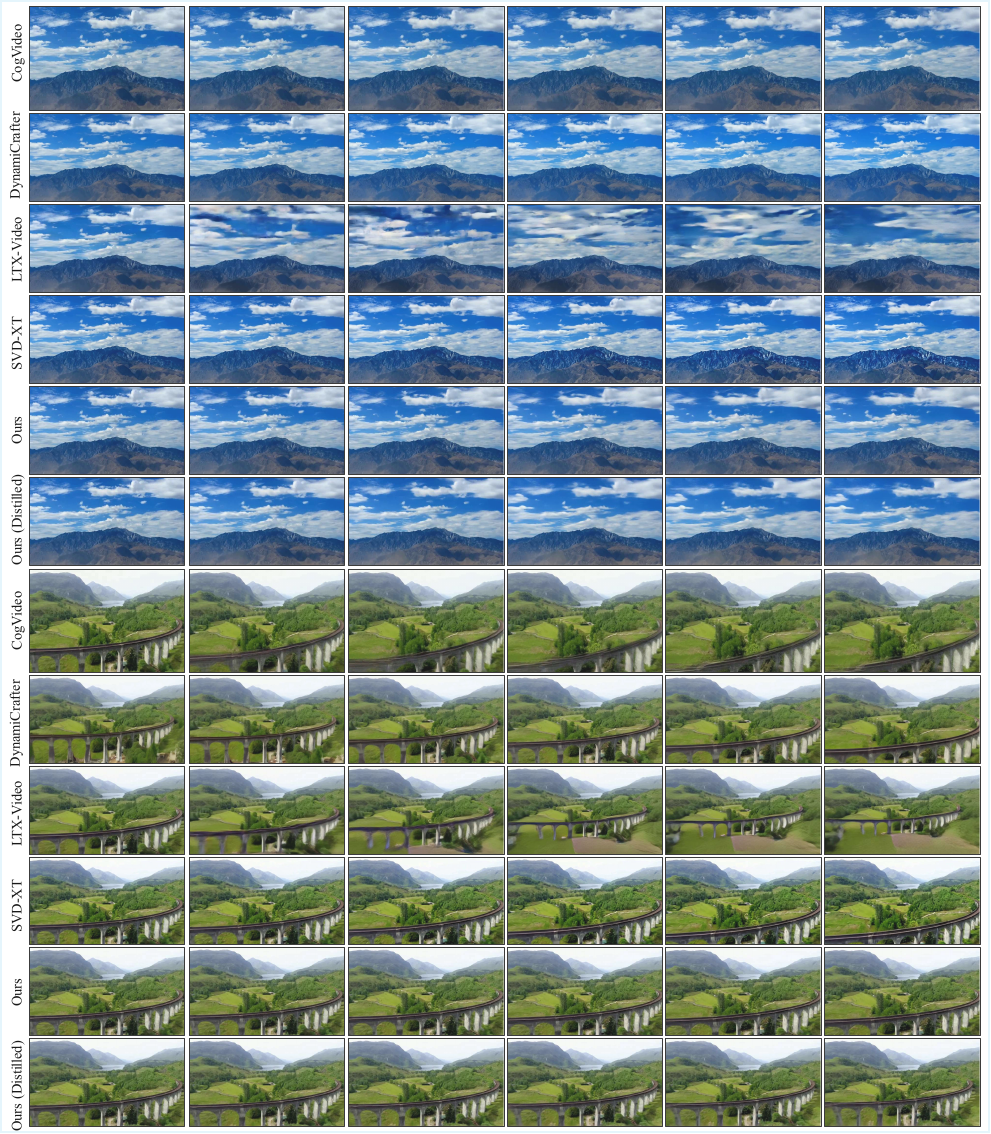}
	\caption{More visual comparison results.} 
	\label{fig:more_vis2}
	\vspace{-10pt}
\end{figure*}

We tested the VbenchI2V indicators of different models, and the results are shown in Tab.~\ref{model_comparasion_appendix}. We offer different model versions, including those before and after distillation. It can be observed that adjusting the motion score significantly affects the dynamic degree index, which in turn influences the overall score. For example, when inferring a landscape video, increasing the motion score from 2 to 5 causes the dynamic degree to change from 0.157 to 0.495. The overall indicator improves, while other indicators decrease slightly. Overall, the proposed model can achieve comparable performance metrics to existing models such as DynamiCraft, CogVideoX1.5, SVD, and LTX-Video. We provide further visual comparison results, as shown in Fig.~\ref{fig:more_vis1} and Fig.~\ref{fig:more_vis2}. For methods that require prompts, we provide standardized prompts, such as ``Driving people in the image to move, such as shaking their heads, talking or smiling.", ``Move the image perspective and convert it into a video."

\begin{table*}[h!]
	\centering
	\scalebox{1}{
	\begin{tabular}{cccccccccccc}
		\toprule
		Model  & {SC} & {BC}  &{MS}& {AQ} & {IQ} & {DD}& Q$_{score}$& {i2v$_{sub}$} & {i2v$_{back}$} & i2v$_{score}$& T$_{score}$\\
		\midrule
		DynamiCrafter & 0.985 &0.979 & 0.988&0.603 &0.695 & 0.172&  0.783&0.987&0.987&0.937&0.860 \\
        CogVideoX1.5 & 0.964 &0.953 & 0.989&0.559 &0.674 & 0.370&  0.779&0.982&0.988&0.935&0.857 \\
		SVD  & 0.986 &0.975 & 0.992&0.557 &0.699 & 0.136&  0.774&0.983&0.988 &0.935&0.855\\ 
		LTX-Video& 0.983 &0.979 & 0.994&0.542 &0.709 & 0.509&0.808 &  0.985&0.988&0.936&0.872  \\
		Ours(U.2.5) & 0.984 &0.971 & 0.993&0.550 &0.710 & 0.264&0.785&  0.989&0.989 &0.940&0.862 \\
		Ours(D.2.2)& 0.989 &0.982 & 0.994&0.552 &0.703 & 0.157& 0.779&  0.986&0.985&  0.935&0.857  \\
		Ours(D.2.5) & 0.984 &0.978 & 0.993&0.554 &0.695 & 0.495&0.808&  0.984&0.983 &0.933&0.869 \\
		\bottomrule
	\end{tabular}
}
\caption{ Comparison of various models generate 17 frame video. (SC: subject consistency
	, BC: background consistency, MS: motion smoothness, AQ: aesthetic quality, IQ: imaging quality, DD: dynamic degree, Q$_{score}$: Quality Score, i2v$_{sub}$: i2v subject, i2v$_{back}$: i2v background, i2v$_{score}$: I2V Score, T$_{score}$: Total Score. U: 30-step results without time-step distillation. D: two-step results after time-step distillation. 2.2/2.5: motion score for human data and scenery data.)}
\label{model_comparasion_appendix}
\end{table*}

\begin{figure*}[h]
	\begin{center}
		{
			\begin{minipage}{8.5cm}
				\centering
				\includegraphics[width=1\textwidth]{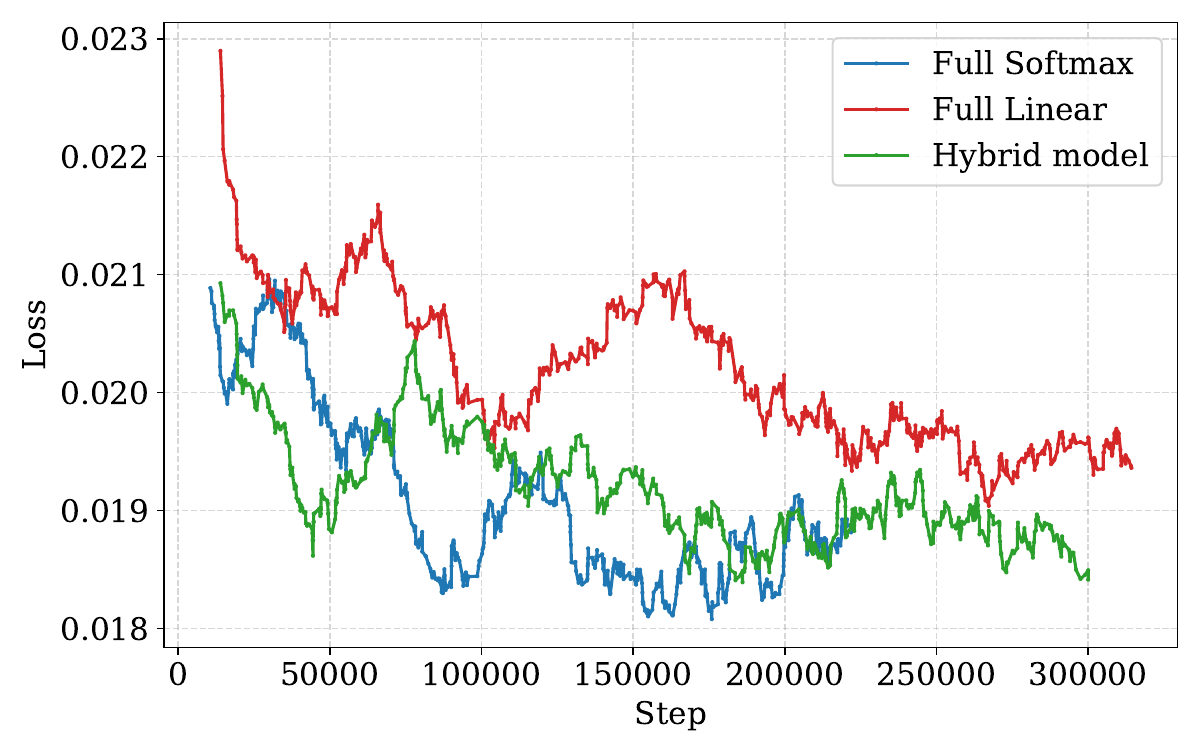}
			\end{minipage}
		}
		{
			\begin{minipage}{8.5cm}
				\centering
				\includegraphics[width=1\textwidth]{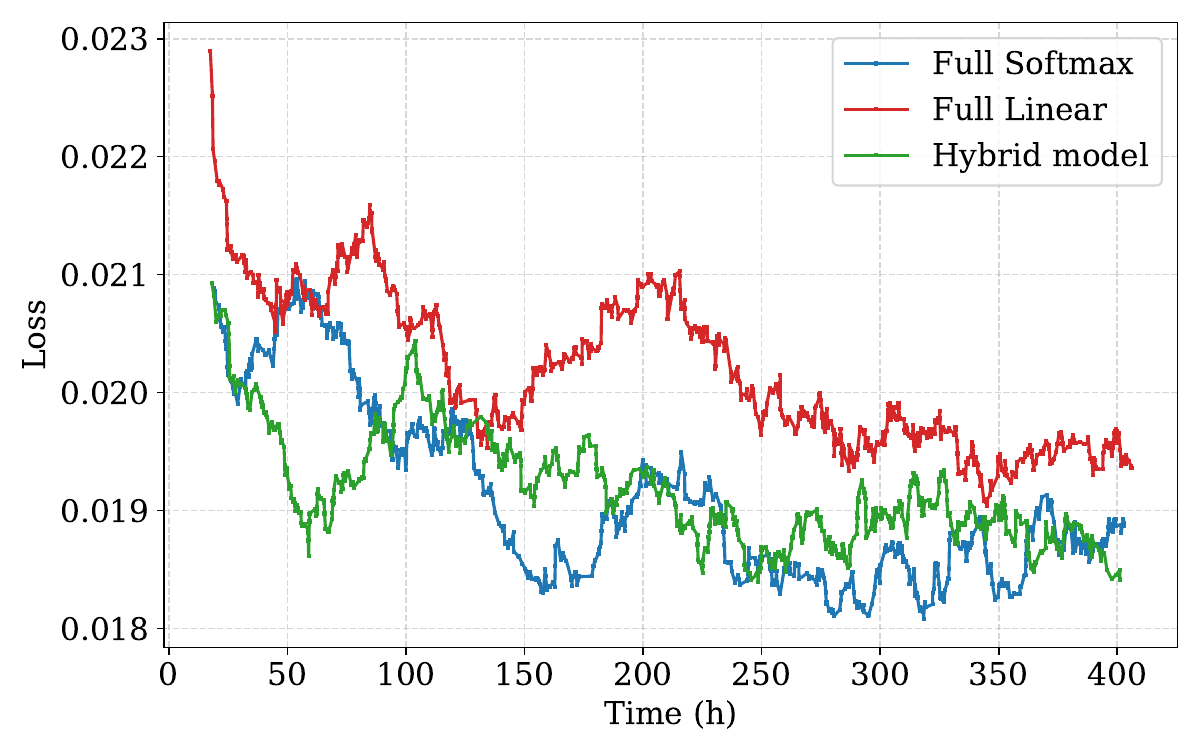}
			\end{minipage}
		}
	\end{center}
	\vspace{-10pt}
	\caption{Training step number and loss curves and training time and loss curves for different types of attention modules.} 
	\label{fig:4}  
\end{figure*}

\begin{figure*}[th]
	\centering
	\includegraphics[width=1\textwidth]{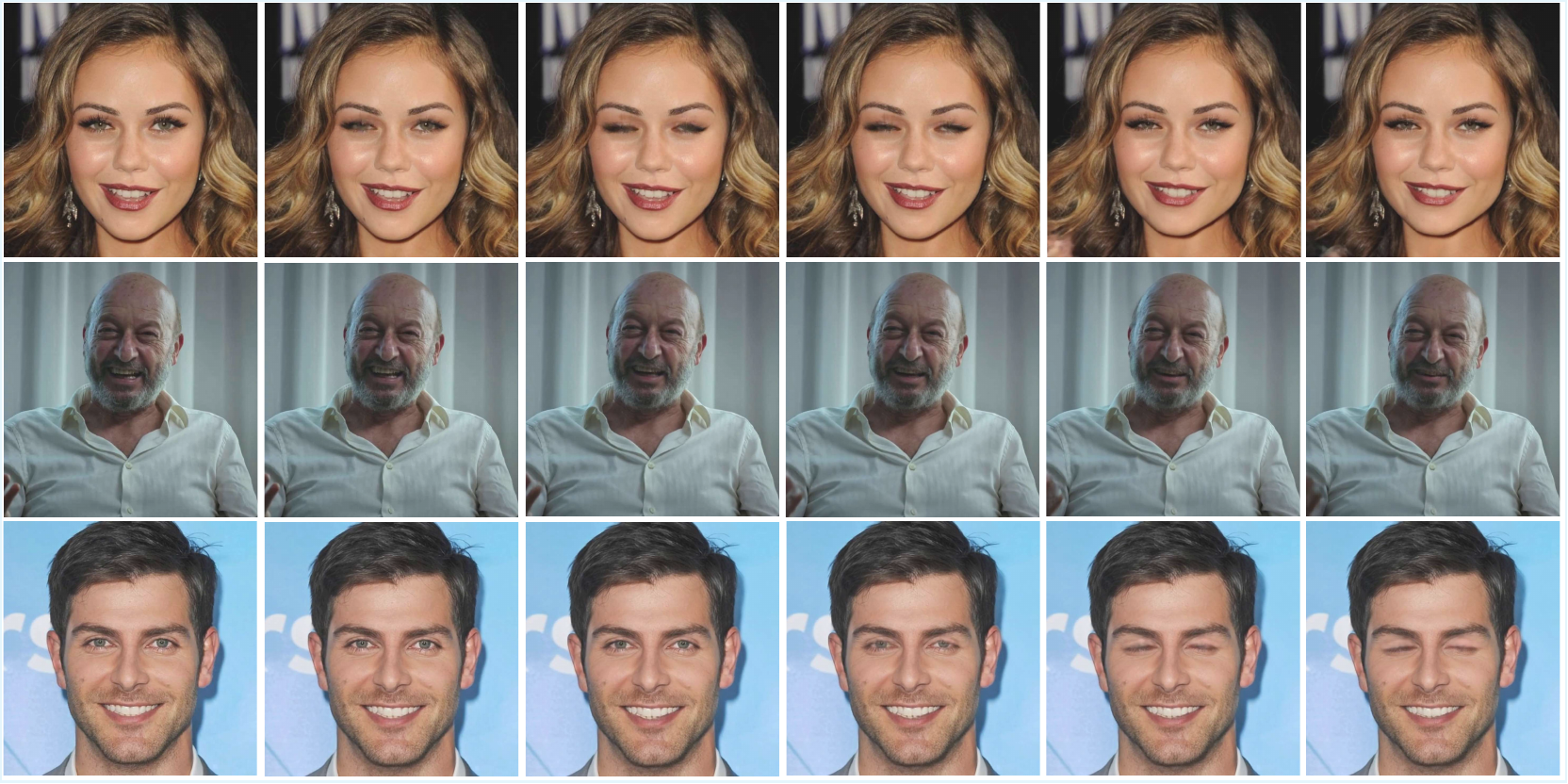}
	\caption{17 frame 960$\times$960 Image to Video Visualization Results (Inference steps: 28).}
	\label{fig:result}
	\vspace{-5pt}
\end{figure*}

\subsection{More Resolution Results}

We trained a model with a 960$\times$960 resolution version on the facial video dataset. The visualization results are shown in Fig.~\ref{fig:result}. From the figure, it can be observed that at this resolution, the model is capable of effectively driving facial expressions, enabling basic actions such as blinking and opening the mouth. Compared with 720P data, the 960-resolution training video content is cleaner and has better generation effect.

\subsection{Ablation Experiment Supplement}
In order to verify the effectiveness of the proposed hybrid attention model, we provide the training loss curves of different types of attention mechanisms, as shown in the Fig.~\ref{fig:4}. As can be seen from the figure, with the same number of training steps and training time, both the full softmax model and the hybrid model significantly outperform the full linear model. At the same training time, the full softmax model is significantly slower than both the full linear model and the hybrid model. This figure demonstrates that the our proposed hybrid model achieves superior results in both speed and performance.

\section{Mobile Results}
\label{s10}

\begin{figure*}[h]
	\centering
	\includegraphics[width=1\linewidth]{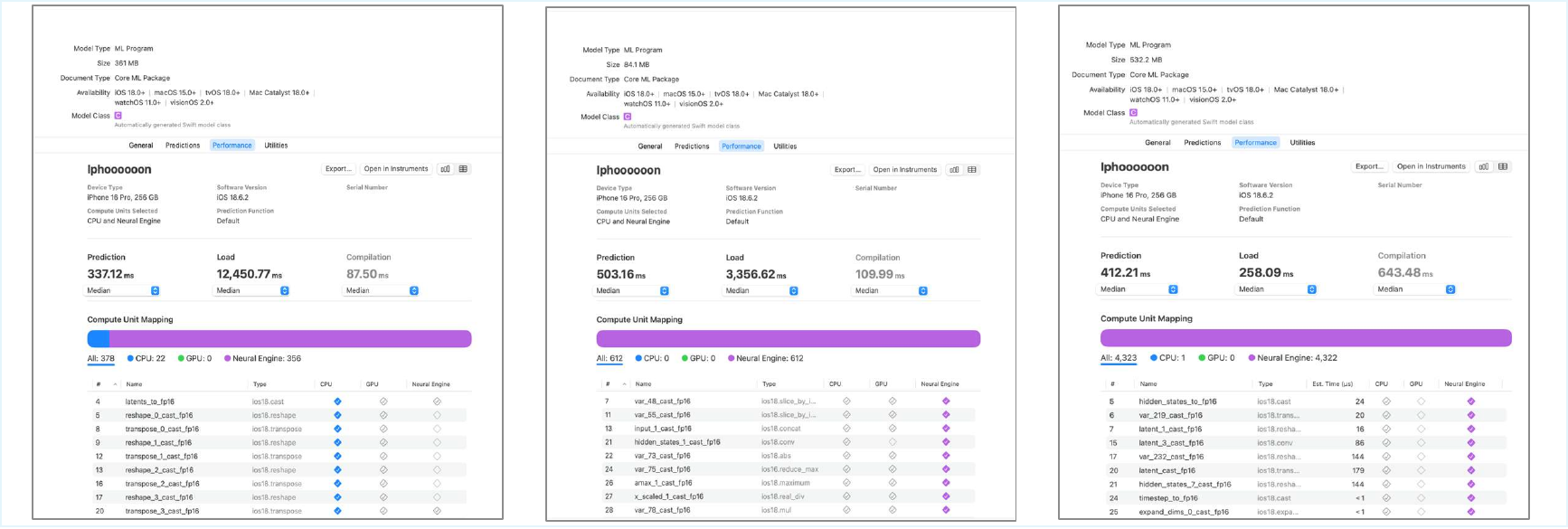}
	\caption{VAE's encoder (left), decoder (middle) and denoiser (right) runtime on mobile devices.} 
	\label{fig:appendix_fig1}
	\vspace{-10pt}
\end{figure*}

\textbf{Model conversion and deployment.} We convert PyTorch models to Core~ML for on-device inference on the iPhone~16~Pro. Specifically, we first export the model to TorchScript and then use \texttt{coremltools} to generate a Core~ML model, which we package in the \texttt{.mlpackage} format. All automatic optimization passes provided by \texttt{coremltools} are enabled (e.g., fusing \texttt{batch\_norm} into \texttt{conv} or \texttt{conv\_transpose}). During inference, FP16 is used for both activations and weights by default, while numerically sensitive operations—such as selected \texttt{RMSNorm} layers—are kept in FP32. To ensure functional equivalence before and after conversion, we compare model outputs on identical inputs and report the peak signal-to-noise ratio (PSNR) between the PyTorch and Core~ML results. All benchmarks are obtained via the Core~ML inference APIs on-device. We tested the speed of the model on mobile devices, as shown in Fig.~\ref{fig:appendix_fig1}.

\noindent
\textbf{Mobile speed.} We tested the speed of the model on mobile devices, as shown in Fig.~\ref{fig:appendix_fig1}. To generate a 1280$\times$720$\times$17 frame video, the inference times for the VAE encoder, decoder, and denoiser on a mobile devices are 337ms, 503ms, and 412ms, respectively. In actual I2V end-to-end testing, the module's time will be approximately 10\% slower compared to when it is tested individually.

\begin{figure*}[h]
	\centering
	\includegraphics[width=1\linewidth]{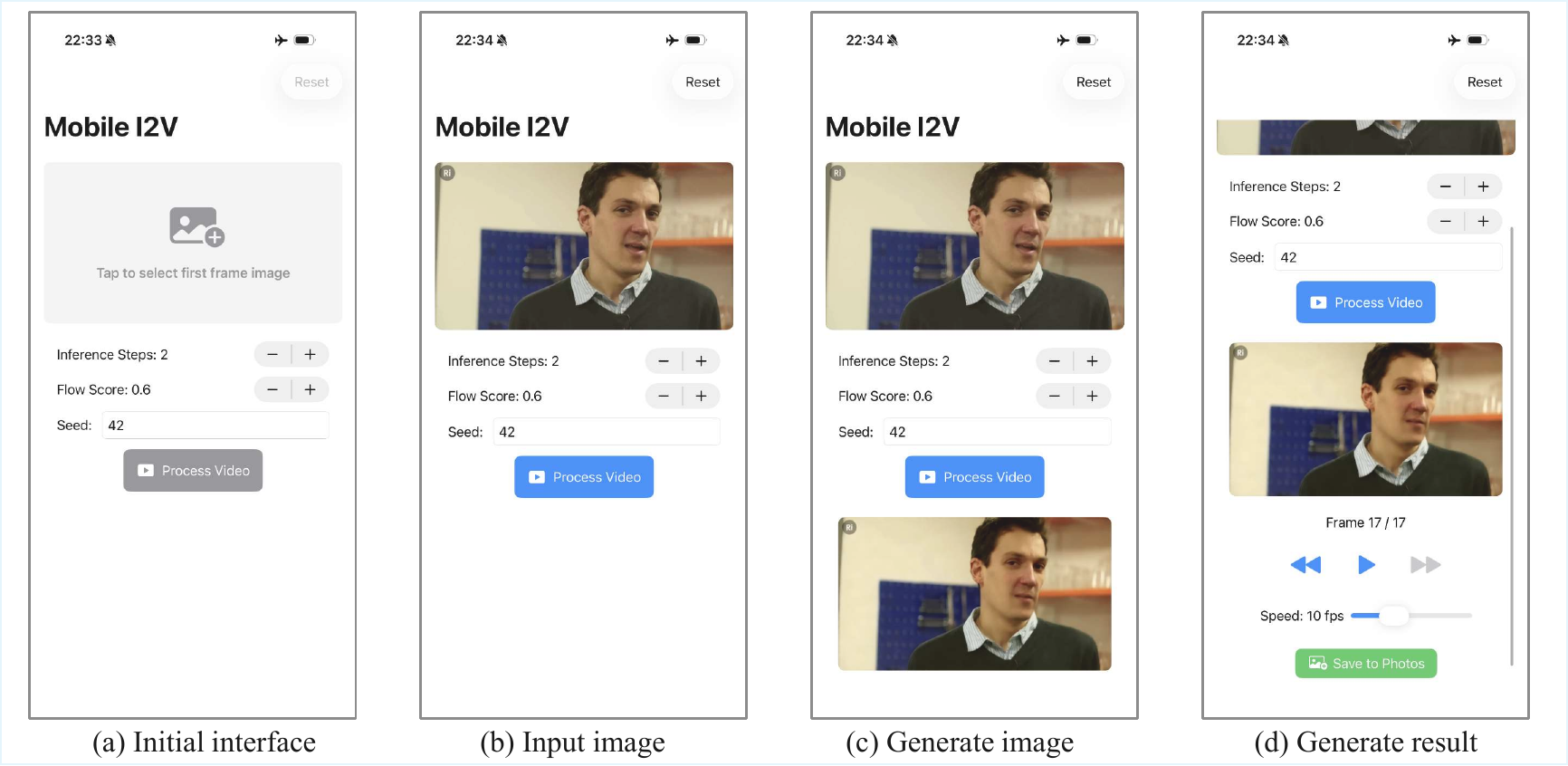}
	\caption{Screenshot of the UI interface on mobile devices.} 
	\label{fig:appendix_fig2}
	\vspace{-10pt}
\end{figure*}

\noindent
\textbf{Mobile UI.} We designed the mobile image to video UI interface to facilitate the effect display, as shown in Fig.~\ref{fig:appendix_fig2}. Using this application, you can read images from your phone and then convert them into videos. The entire I2V process is carried out on the mobile devicce.

\section{Limitations and Future Work}
\label{s11}
\begin{figure*}[h]
	\centering
	\includegraphics[width=1\linewidth]{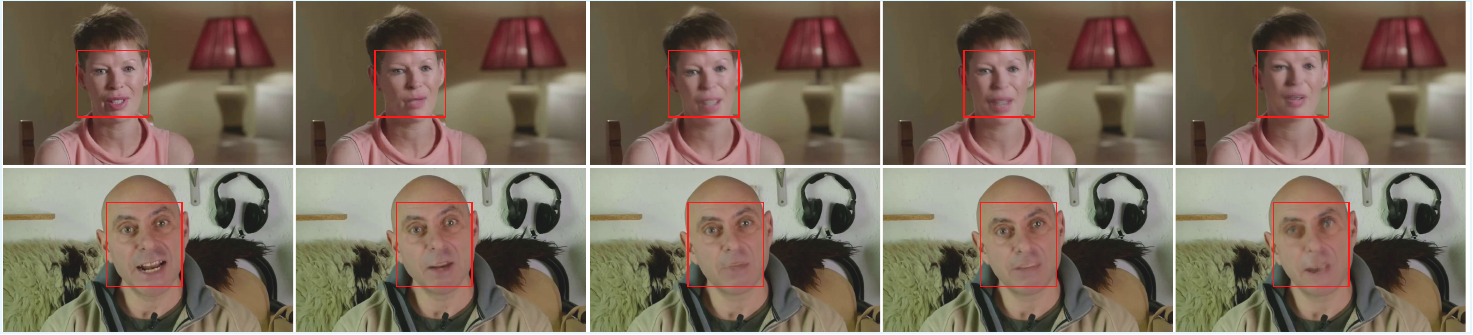}
	\caption{Some generated failure cases.} 
\label{fig:failure_case}
\vspace{-10pt}
\end{figure*}

\textbf{Limitations.} Since we use a 32$\times$32$\times$8 high compression rate VAE, blurring may occur in complex dynamic scenes such as faces. As shown in the Fig.~\ref{fig:failure_case}, complex dynamic areas such as the mouth and eyes of a person's face are prone to blurring. In addition, due to limited data resources, our model is mainly trained on faces and aerial scenery, resulting in poor performance on other scenes.

\noindent
\textbf{Future work.} From the perspective of model architecture, a high compression VAE is crucial for accelerating generation. Currently, the capabilities of a 32$\times$32$\times$8 compression VAE need to be improved. In addition, in this paper we have seen the potential of linear architecture on mobile devices, and adopting a more powerful linear architecture may be a means to improve the effect. 

\end{document}